\documentclass[letterpaper, 10 pt, conference]{ieeeconf}  % Comment this line out if you need a4paper

\usepackage{booktabs} % For professional looking tables
\usepackage{tabularx} % Allows having table columns that auto-size
\usepackage{xcolor}
\usepackage{threeparttable} % 用于添加表格注释
\usepackage{amsmath}
\usepackage{pifont}
\usepackage{cite}

\usepackage{dblfloatfix}
\usepackage{blindtext}
\usepackage{listings}

\usepackage{graphicx}
\usepackage{listings}
\IEEEoverridecommandlockouts                              % This command is only needed if 
\title{\LARGE \bf
VisFly: An Efficient and Versatile Simulator for Training Vision-based Flight
}

\author{Fanxing Li, Fangyu Sun, Tianbao Zhang, Danping Zou$^*$ 
}

\newcommand{\reffig}[1]{Figure \ref{#1}}
\newcommand{\reftab}[1]{Table \ref{#1}}

\newcommand{\refsec}[1]{Section \ref{#1}}

\begin{document}

\maketitle
\thispagestyle{empty}
\pagestyle{empty}

%%%%%%%%%%%%%%%%%%%%%%%%%%%%%%%%%%%%%%%%%%%%%%%%%%%%%%%%%%%%%%%%%%%%%%%%%%%%%%%%
\begin{abstract}

We present VisFly, a quadrotor simulator designed to efficiently train vision-based flight policies using reinforcement learning algorithms. VisFly offers a user-friendly framework and interfaces, leveraging Habitat-Sim's rendering engines to achieve frame rates exceeding 10,000 frames per second for rendering motion and sensor data. The simulator incorporates differentiable physics and is seamlessly wrapped with the Gym environment, facilitating the straightforward implementation of various learning algorithms. It supports the directly importing open-source scene datasets compatible with Habitat-Sim, enabling training on diverse real-world environments simultaneously. To validate our simulator, we also make three reinforcement learning examples for typical flight tasks relying on visual observations. The simulator is now available at [https://github.com/SJTU-ViSYS-team/VisFly].

\end{abstract}

%%%%%%%%%%%%%%%%%%%%%%%%%%%%%%%%%%%%%%%%%%%%%%%%%%%%%%%%%%%%%%%%%%%%%%%%%%%%%%%%

\section{Introduction}
Quadrotors have great potential applications in various tasks such as search-and-rescue, surveillance, and inspection operations. The capability of autonomous flight of quadrotors is the key to enabling them to accomplish complex tasks. Typically, an autonomous flight stack consists of localization, environmental perception, planning, and control components. However, the classic pipeline heavily relies on the performance of individual components. The accumulation of errors and running time of individual components could greatly reduce the agility and robustness of autonomous flight \cite{RN27}. 

An alternative approach is to train a neural network to process sensory input and directly control the quadrotor. Reinforcement learning (RL) has shown great success in various fields \cite{RLAutoDriving, StarCraft, Humanoid} including aerial robotics \cite{kalakrishnan2011learning, song2023reaching, kaufmann2023champion, RN27}, relying on simulators for large-scale data generation. While real-world data collection is costly and inefficient \cite{RN29, RN15,RN16}, simulation-generated datasets can be used to train policies that could be deployed in real world \cite{tobin2017domain}. Existing simulators \cite{Gazebo,Airsim, Rotors,CrazyS}, though widely used, often suffer from low sampling efficiency, particularly for vision-based tasks. Some quadrotor simulators \cite{flightmare, OmniDrone, pybullet} have addressed these drawbacks by improving the sampling of physical state data, but they still struggle with high-frame-rate vision input. Given its rich and cost-effective sources of perception, vision holds significant potential for exploring high-level autonomous flight in real-world applications.

To handle these issues, we developed an efficient simulation environment for training vision-based autonomous flight of quadrotors. Our environment leverages the PyTorch, Habitat-Sim \cite{puig2023habitat}, and OpenAI's Gym \cite{brockman2016openai} platforms. We implemented flight dynamics and four widely used flight controllers using PyTorch, making flight physics not only computationally efficient but also differentiable. By adopting Habitat-Sim's rendering engine, our environment can achieve high frame rates at 10,000 frames per second while rendering color images at a resolution of 64×64. Moreover, our simulator can load a series of open-source datasets comprising a vast number of 3D scene models (including real-world 3D scans), allowing us to train vision-based flight policies in diverse and realistic scenes. Our simulator is highly flexible, featuring gym-wrapped interfaces for integrating the latest learning algorithms and customized tasks.
To validate our environment, we present three examples of using a general network architecture to train vision-based flight policies for three autonomous flight tasks: landing, navigation, and cooperative flight through the narrow gap. The results demonstrate that our simulation environment can be used to train vision-based flight policies in both single and multi-agent settings with high success rates in a very short training time.  

% To highlight the novel features of our simulator, we compare it with existing simulators in \reftab{tab:comparision}. The contributions of this work could be summarized as:
%  \begin{enumerate}
%     \item We present a novel quadrotor simulator that achieves high performance in rendering both physics and vision data, allowing to train vision-based flight efficiently.
%     \item Our simulator supports single-agent and swarm scenarios in parallelized physics and scenes, with Gym-wrapped interfaces for implementing the latest learning algorithms and customization of tasks. It also integrates open-source datasets for diverse scenes.
%     \item We demonstrate the effectiveness of our training environment for training diverse vision-based tasks, highlighting its versatility. 
% \end{enumerate}

% \FloatBarrier
% \FloatBarrier

\begin{table*}[ht!]
\scriptsize
\centering
\caption{Key Character Comparison of Popular Simulators\label{tab:comparision}}
\label{tab:your-table}
\begin{tabularx}{\textwidth}{l | l l | c c | c c | c| c |c |c|c}
\toprule
& \multicolumn{2}{c|}{\textbf{Engine}} & \multicolumn{2}{c|}{\textbf{Parallel}} &  \multicolumn{2}{c|}{\textbf{FPS}} &\textbf{Sensor}&\textbf{Gym-} & \textbf{3D } &\textbf{Different-} &\textbf{Swarm} \\
% & \multicolumn{2}{c|}{\textbf{Engine}} & & & & \multicolumn{2}{c|}{\textbf{Rapidity}} & \\

& \textbf{Physics} & \textbf{Render} & \textbf{agent} & \textbf{scene} & \textbf{Physics} & \textbf{Render} &  & \textbf{wrap} & \textbf{Datasets} &\textbf{iable} & \\
\midrule
\textbf{RotorS$^*$}\cite{Rotors} & Gazebo & OpenGL & \ding{55} & \ding{55} & $1 \times 10^2$ & 20 & I, D, C & \ding{55} & \ding{55} & \ding{55} & \ding{55} \\
\textbf{Airsim$^*$}\cite{Airsim} & PhysX & Unreal & \ding{51} & \ding{55} & $1 \times 10^2$ & 60 & I, D, C, S & \ding{55} & \ding{55} & \ding{55} & \ding{51} \\
\textbf{FlightGoggles$^*$}\cite{FlightGoggles} & Flexible & Unity & \ding{55} & \ding{55} & $1 \times 10^3$ & 64 & I, D, C, S & \ding{55} & \ding{55} & \ding{55} & \ding{55} \\
\textbf{FastSim$^*$ }\cite{FastSim} & Flexible & Unity & \ding{51} & \ding{55} & $1 \times 10^2$ & 40 & I, D, C, S & \ding{55} & \ding{55} & \ding{55} & \ding{51} \\
\textbf{CrazyS$^*$ }\cite{CrazyS} & Gazebo & OpenGL & \ding{55} & \ding{55} & $1 \times 10^3$ & 20 & I, D, C & \ding{55} & \ding{55} & \ding{55} & \ding{55} \\
\textbf{Flightmare}\cite{flightmare} & Flexible & Unity & \ding{51} & \ding{55} & $\mathbf{2.2 \times 10^5}$ & $2.3 \times 10^2$ & I, D, C, S & \ding{51} & \ding{55} & \ding{55} & \ding{55} \\
\textbf{PyBulletDrone}\cite{pybullet} & Bullet & OpenGL & \ding{51} & \ding{55} & $1 \times 10^4$ & $3 \times 10^3$ & I, D, C, S & \ding{51} & \ding{55} & \ding{55} & \ding{51} \\
\textbf{OmniDrone}\cite{OmniDrone} & Self-defined & Isaac Sim & \ding{51} & \ding{55} & $2 \times 10^5$ & $3 \times 10^2$ & I, D, C, S, L & \ding{51} & \ding{55} & \ding{55} & \ding{51} \\
\midrule
\textbf{Ours} & Self-defined & Habitat-sim & \ding{51} & \ding{51} & $3.7 \times 10^4$ & $\mathbf{1 \times 10^4}$ & I, D, C, S & \ding{51} & \ding{51} & \ding{51} & \ding{51} \\
\bottomrule
\end{tabularx}

\begin{tablenotes}
\item[1] I, C, D, S, and L refer to IMU, RGB, depth, segmentation, and LiDAR, respectively. Simulators marked with an asterisk (*) are full-stack, while the others are learning-specialized. VisFly is lightweight and does not have high hardware requirements. Note that the FPS data provided by the original papers were not evaluated with the same number of agents, but they still offer a performance reference. All frame rates are tested with 100 environments if parallelization is supported, and render FPS are tested while rendering $64\times64$ depth images. FastSim cannot be deployed locally at present, so the FPS data is directly cited from the original paper. All frame rates are tested on a desktop with an RTX 4090 and a 32-core 13th Gen Intel(R) Core(TM) i9-13900K.
\end{tablenotes}

\end{table*}
\section{Related Work}
% two steps to introduce current simulators

\subsection{Full-stack simulators}
We compare the key factors of primary simulators in \reftab{tab:comparision}.
The initial aim of simulators is to create a simulated platform for users to verify the algorithmic prototypes because quadrotors with immature algorithms are easy to crash. This category always consists of comprehensive full-stack simulated flight modules such as perception, planning, and control, especially for highly realistic physics. Gazebo \cite{Gazebo} is a widely used full-stack robotics simulator of this category. While it is a general-purpose simulator, it strongly supports UAVs through various plugins and models.  
Considering various requirements, researchers develop simulators specific for UAVs based on Gazebo. 
RotorS \cite{Rotors} is a popular multi-rotor UAV modular simulator built on Gazebo,  providing flexible interfaces to replace several parts of the pipeline, making it convenient to verify state estimators or control algorithms. 
CrazyS \cite{CrazyS} is a lightweight simulator specialized for a software-to-hardware solution Crazyflie 2.0, which helps to deploy in the real world easily.

Rather than using open-source rendering engines as in Gazebo, several simulators utilize commercial rendering engines such as Unity\footnote{https://unity.com/} and Unreal Engine\footnote{https://www.unrealengine.com/} to achieve high visual fidelity. FlightGoogles\cite{FlightGoggles} employs the Unity engine for realistically simulating image sensors and uses real-world flight data to create avatars in simulation environments, enabling vehicle-in-the-loop simulation. Similarly, AirSim \cite{Airsim} is built on Unreal Engine to offer physically and visually realistic environments, along with real-time hardware-in-the-loop (HITL) simulation capabilities. Additionally, AvoidBench \cite{yu2023avoidbench}, built on the Unity engine, offers valuable benchmarks to evaluate obstacle avoidance algorithms. FastSim \cite{FastSim}, also built on Unity,  stands out as the first to integrate path planning algorithms directly into the simulator, facilitating tests of high-level autonomous control functionalities. 
While using commercial engines significantly enhances image quality (including depth maps and semantic segmentation), it comes at the cost of slower rendering speed because of their sophisticated rendering pipelines, particularly for the vision data. However, for applications focusing on learning in simulators using vision sensors, it is critical to acquire  vision data quickly for fast training. 

\subsection{Learning-specialized simulators}
High acquisition rates for both sensory data and robot physics significantly accelerate the training process of learning-based algorithms. Notably, physics computation can be achieved at very high frequencies in some existing simulators. For instance, by paralleled running, the frame rate of physics computation can reach an impressive 100000 FPS in Flightmare \cite{flightmare} and OmniDrone \cite{OmniDrone}. However, the frame rates will drop to hundreds of FPS if the visual observation is rendered simultaneously in Flightmare. 
PyBulletDrone \cite{pybullet} achieves high rendering FPS via simple rendering engine OpenGL3 and TinyRender\footnote{https://github.com/ssloy/tinyrenderer}. However, the direct basic OpenGL-based rendering engine lacks the functionality to import complex scenes and objects, making it difficult to leverage open-source scene datasets. 

Selecting a rendering platform for training vision-based flight policies requires balancing speed, open-source scene availability, and hardware requirements. Habitat-sim, a lightweight simulator, excels in rendering speed and supports popular open-source datasets with minimal hardware demands. However, it primarily focuses on ground robots and manipulations. In contrast, Nvidia's Isaac Sim \footnote{https://developer.nvidia.com/isaac/sim}  is a comprehensive simulator that features extensive joint optimizations in rendering and simulation at the software architecture level. Nevertheless, it has higher hardware requirements and provides less flexibility and extensibility compared to Habitat-sim.
Our simulator, built on Habitat-sim, leverages its fast rendering engine and extensive 3D datasets but is specifically tailored for aerial robots. Unlike other simulators focused on state-only tasks, VisFly emphasizes vision-based flights and fully integrates with OpenAI's Gym standards for easy customization. VisFly also provides modular examples and tutorials to help users quickly start training vision-based UAV tasks.

% Our simulator is built on Habitat-sim, utilizing its fast rendering engine and well-designed interfaces to leverage existing massive 3D datasets. However, unlike Habitat-sim, which focuses on ground robots and manipulations, our simulator is tailored specifically for aerial robots. For the convenience of implementation, existing learning-specialized simulators, such as Flightmare, OmniDrone, and PyBulletDrone wrap the environment upon OpenAI's gym environment standards. An OpenAI's \emph{gym} class receives an action, updates its internal state, and returns a new state paired with a corresponding reward. 
% Like most of existing simulators, VisFly provides examples and tutorials to help beginners get started quickly. However, unlike others that primarily focus on state-only tasks—using only the UAV's state, such as orientation, position, and velocities for control or planning—VisFly's examples are specifically designed for vision-based flights. While some simulators, like OmniDrone, include examples using LiDAR inputs, VisFly emphasizes using visual data. Our examples are released with a modular design and are seamlessly integrated with Gym environments, providing users with an easy way to customize their vision-based flight tasks.

\begin{figure*}[t!]
\centering
\includegraphics[width=\textwidth]{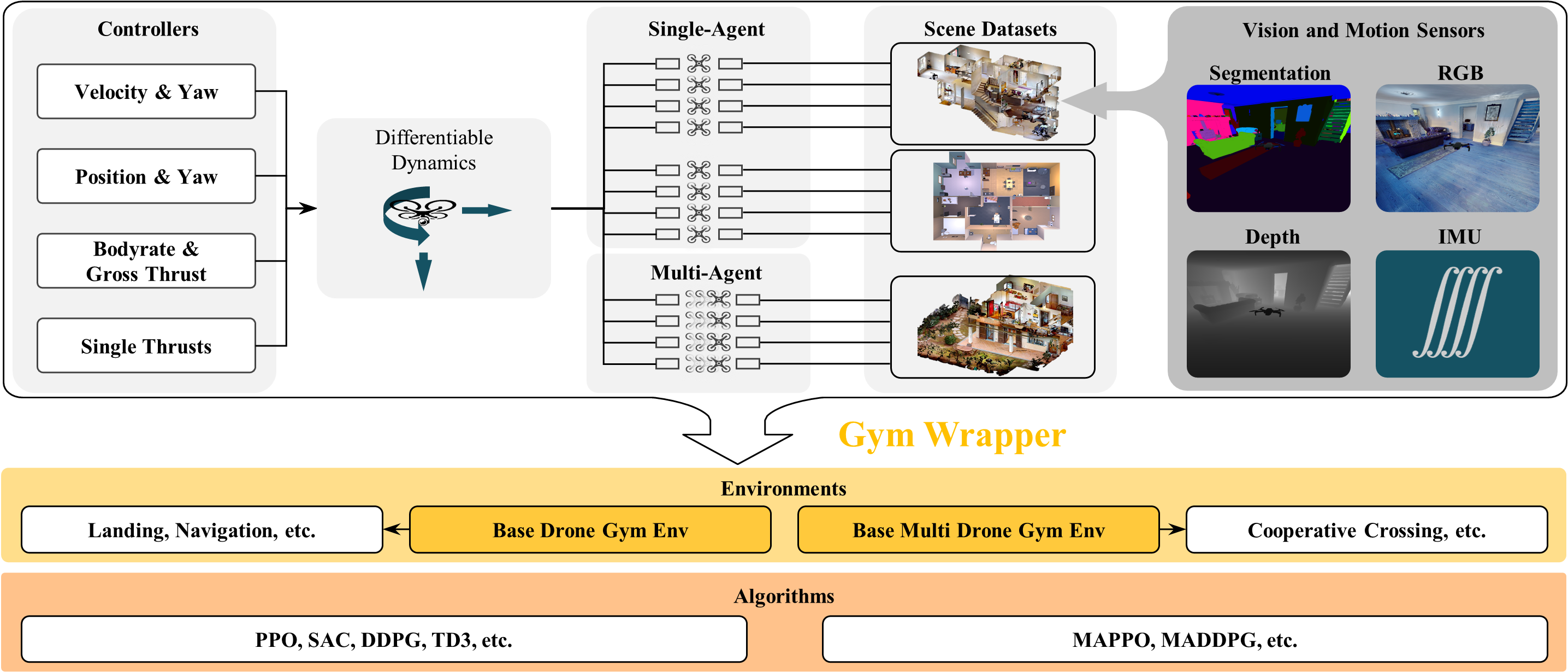}
\caption{Overall diagram of VisFly. Our system uses differentiable physics to drive single or multiple quadrotor agents through four different types of controllers and employs Habitat-sim’s rendering engine for fast rendering and access to open-source datasets. All of these components are integrated into Gym environments, providing standard interfaces for various learning algorithms.}
\label{fig:overview}
\end{figure*}

\section{Methodology}
Here, we introduce three key features of VisFly and how they are implemented: (1) differentiable, and parallelized dynamics computation; (2) a rapid-rendering, parallelized, photo-realistic scene manager;
(3) Gym-wrapped interfaces and extensions based on modular design. A schematic diagram of VisFly is shown in \reffig{fig:overview}. %

\begin{figure*}
    \centering
    \includegraphics[width=\linewidth]{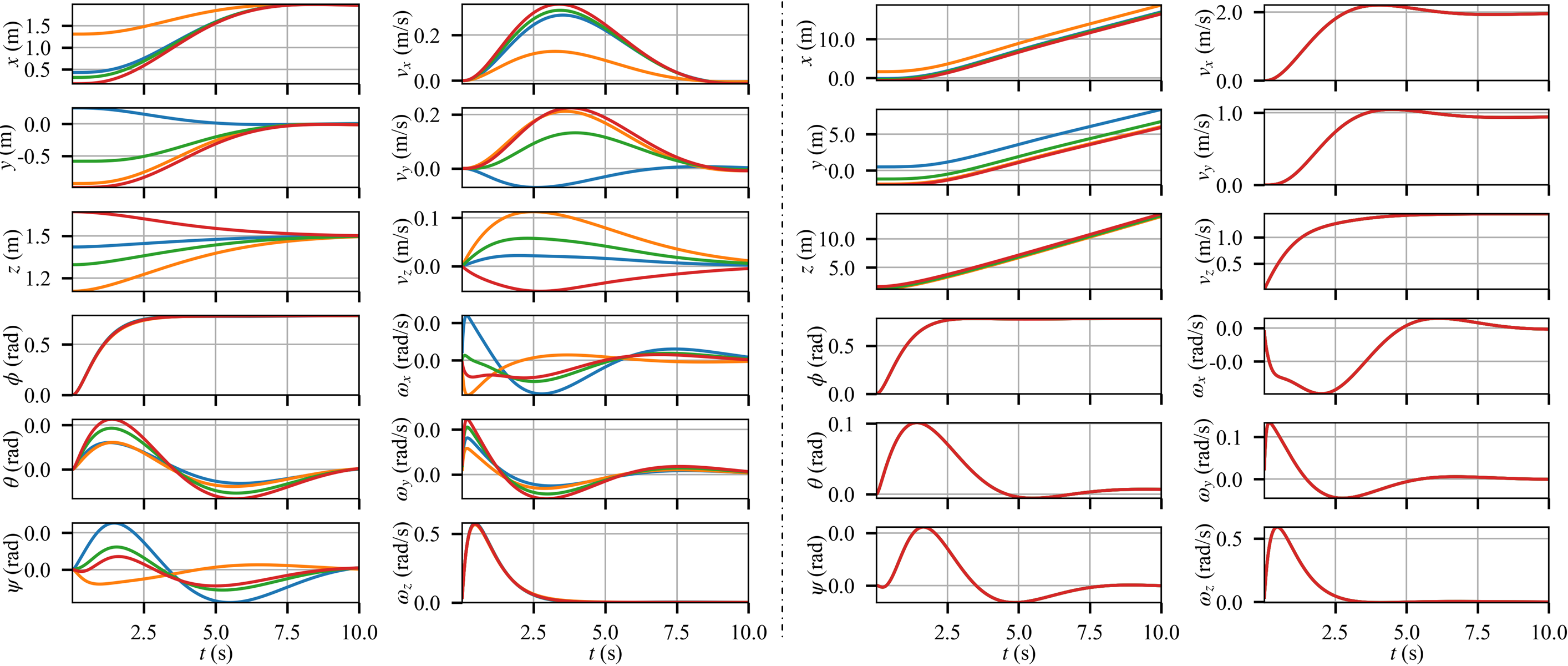}
    \caption{Response Curves of Position, Orientation, Linear Velocity, and Angular Velocity to Step Signals of \textbf{Left}:position commands and \textbf{Right}:linear velocity commands. Different colors denote various initial states.}
    \label{fig:ctrl}
\end{figure*}

\begin{figure*}[h!]
\centering
\includegraphics[width=\textwidth]{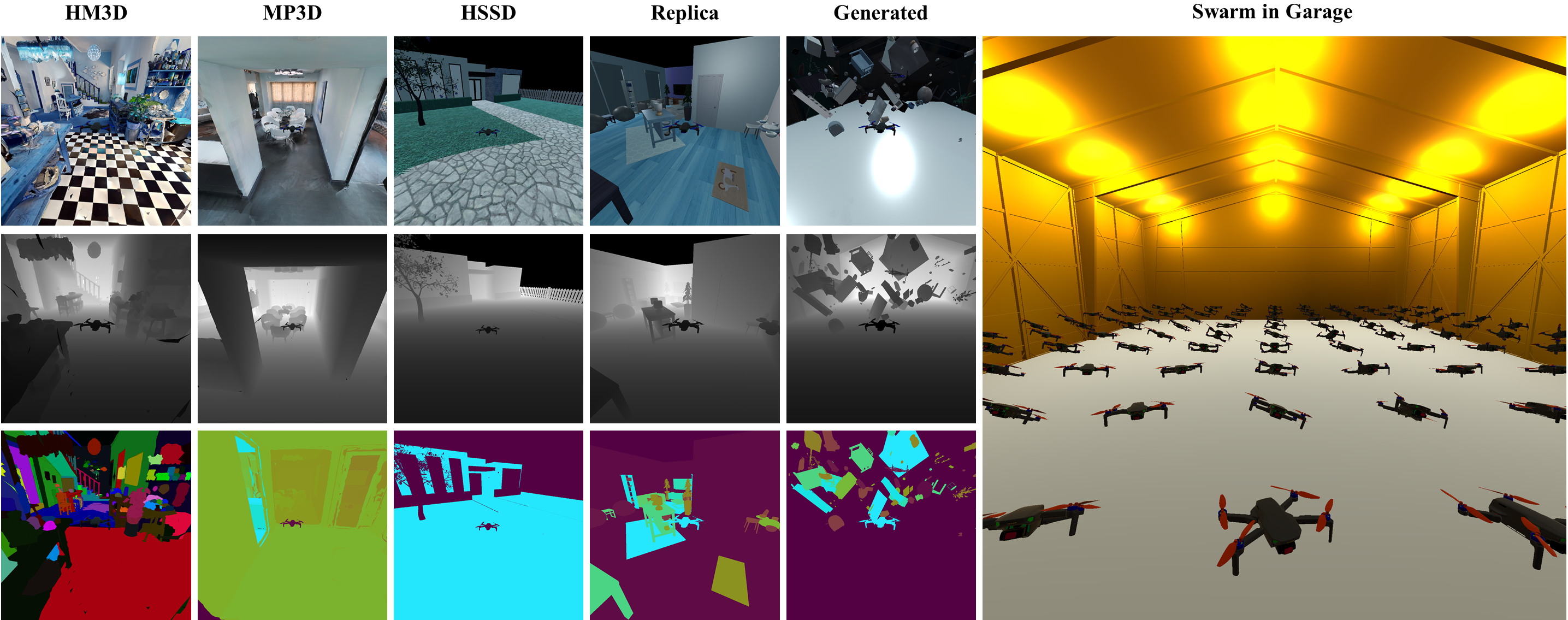}
\caption{RGB, depth, and semantic view of agent in open-source and customized scenes. The right sub-image illustrates a swarm of 100 quadrotors in a clear garage. }
\label{fig:render}
\end{figure*}

\subsection{Differentiable dynamics based on PyTorch}
The dynamics simulation is implemented using PyTorch to enable efficient, parallel physics computations on GPUs. PyTorch's computational graph also enables the computation of analytical gradients through back-propagation, making the dynamics computation differentiable.  The differentiable dynamic model can be directly integrated into model-based reinforcement learning frameworks\cite{moerland2023model}.  Analytical gradients computed by differentiable dynamics offer more precise descent directions, significantly accelerating training compared to common value-based or policy-based reinforcement learning methods\cite{wiedemann2023training,song2024learning}. 
We model the quadrotor's dynamics as follows: 
\begin{equation}
\small
\label{eq:dynamics}
\begin{aligned}
    &\dot{\mathbf{x}}_W = {\mathbf{v}}_W \\
    &\dot{\mathbf{v}}_W = \frac{1}{m} \mathbf{R}_{WB}(\mathbf{f} + \mathbf{d}) + \mathbf{g} \\
    &\dot{\mathbf{q}} = \frac{1}{2} \mathbf{q} \otimes \mathbf{\Omega} \\
    &\dot{\mathbf{\Omega}} = \mathbf{J}^{-1} (\eta - \mathbf{\Omega} \times \mathbf{J} \mathbf{\Omega})
\end{aligned}
\end{equation}
where $\mathbf{x}_W$, $\mathbf{v}_W$, $\mathbf{q}$ represent the position, velocity, and orientation (quaternion) of quadrotors in the world frame, and $\mathbf{\Omega}$ denotes the angular velocity in the body frame. The quaternion-vector product is represented by $\otimes$. The rotation matrix from the body frame to the world frame is $\mathbf{R}_{WB}$. The collective rotor thrust and the momentum are denoted by $\mathbf{f}$ and $\mathbf{\eta}$, respectively. The diagonal inertia matrix is $\mathbf{J}$, and the quadrotor mass is $m$. The air drag in the body coordinate, assumed proportional to the square of velocity \cite{houghton2003aerodynamics}, is represented by $\mathbf{d}$: 
\begin{equation}
\small
    \mathbf{d}=\frac{1}{2}\rho \mathbf{v}_B \odot \mathbf{v}_B \mathbf{C}_d  \odot \mathbf{s}
\end{equation}
where $\rho$, $\mathbf{C}_d$, $\mathbf{s}$, $\mathbf{v}_B$ denote the air density, drag coefficients, crossing area, and velocity in the body frame, respectively, the element-wise multiplication is denoted by $\odot$. The collective thrust $\mathbf{f}$ and momentum $\mathbf{\eta}$,  are computed from individual rotor thrusts  $\mathbf{f}_i$:
\begin{equation}
\small
\label{eq:f}
    \mathbf{f}=\sum_{1}^{4}\mathbf{f}_i ,\quad 
    \mathbf{\eta}=\sum_{1}^{4}\mathbf{T}_i\times \mathbf{f}_i. 
\end{equation}
Here $\mathbf{T}_i$ denotes the moment arm matrix for the rotor $i$. To account for control input delay, we introduce the time decay constant $c$ and model it as an exponential process:
\begin{equation}
\small
\mathbf{f}_i(3)= k_2 \omega_i^2 + k_1 \omega_i + k_0 ,\quad \omega_i = \omega_i^{des} + (\omega_i'-\omega_i^{des})e^{-ct}
\end{equation}
where $\omega_i$ denotes the spinning speed for rotor $i$, and $\omega_i'$, $\omega_i^{des}$ indicate the current and desired states, respectively, the constants $k_2,k_1,k_0$ are thrust coefficients with respect to rotor speeds. $\mathbf{f}_i(3)$ is the thrust along $z$-axis of rotors. Based on \cite{lee2010control, 7139420}, we develop four widely-used control methods: individual single-rotor thrusts (SRT), mass-normalized collective thrust and body rates (CTBR), position commands and yaw (PS), and linear velocity commands and yaw (LV). SRT and CTBR are control inputs for Betaflight, which can be used for training efficient end-to-end policies\cite{Kaufmann2022Benchmark}. We evaluate the LV and PS controllers with step commands. The responses of states shown in \reffig{fig:ctrl}. The minimal simulation time step can be set several times shorter than the control time step, achieving more precise physics. Both the fourth-order Runge-Kutta and Euler methods are supported, providing users the flexibility to choose according to their platform's performance.

\subsection{Efficient rendering and open-source scene management}
\begin{figure}[h]
\centering
\includegraphics[width=\linewidth]{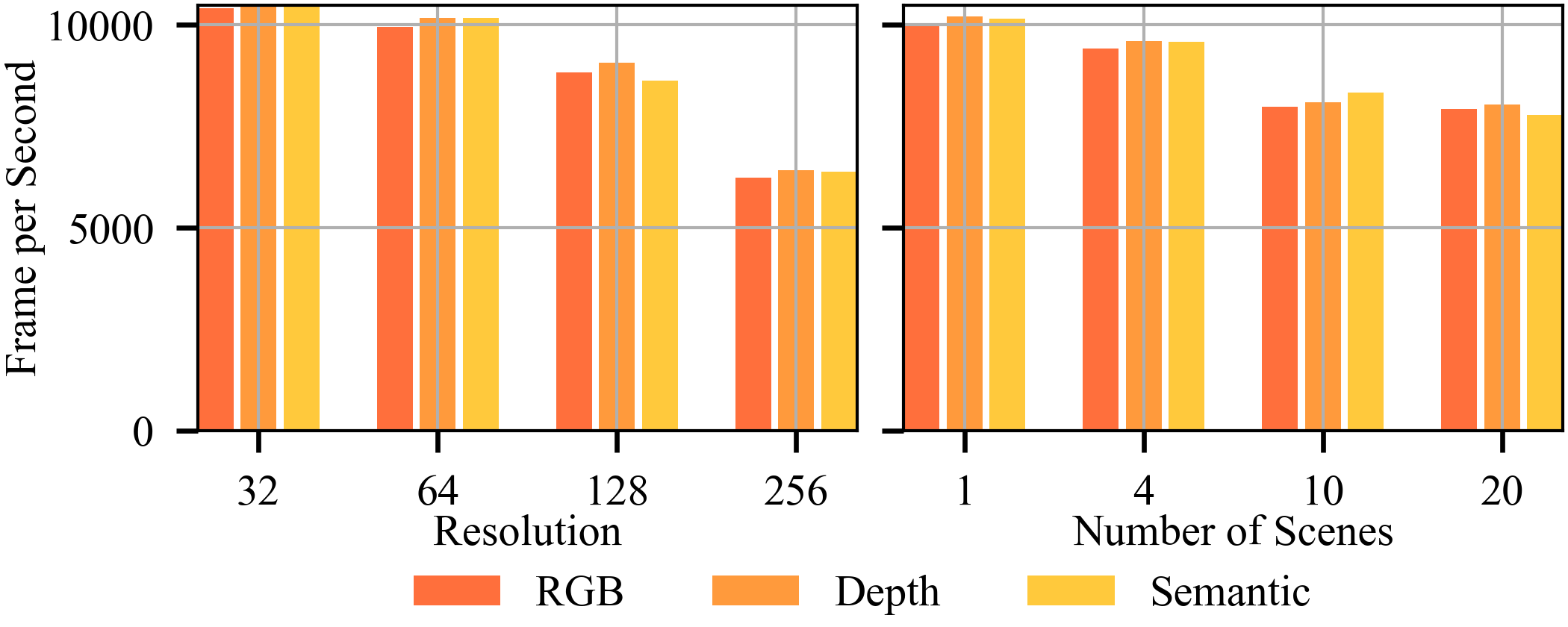}
\caption{
Frame rate performance of VisFly. \textbf{Left}: Tested with 100 agents in the Replica dataset, achieving up to 10,000 FPS. At 256$\times$256 resolution, the frame rate still reaches up to 6,000 FPS. \textbf{Right}: Variation in frame rate with the number of scenes running simultaneously at 64$\times$64 resolution.
}
\label{fig:fps}
\end{figure}

We use the rapid rendering engine of Habitat-sim to render the visual observation data efficiently. \reffig{fig:fps} presents the frame rate of VisFly while running physics simulations. With visual input, VisFly can achieve a frame rate of up to 10,000 FPS, almost 100 times faster than simulators using Unity or Unreal Engine. 
% Compared to other simulators, researchers can train models with different architectures or parameters multiple times within the limited time, significantly increasing iteration efficiency.
Similar to Habitat-Sim, VisFly supports a range of open-source 3D scene datasets, including HM3D\cite{ramakrishnan2021habitat}, MP3D\cite{chang2017matterport3d}, Replica\cite{straub2019replica}, and HSSD\cite{khanna2023habitat}, many featuring highly photorealistic real-world scans. These datasets facilitate direct training and evaluation of models. To augment these resources, we incorporate a scene generator for producing random, cluttered environments, such as those used in collision avoidance training. The combination of real-world data and customizable scenes enhances VisFly's adaptability, reducing Sim-to-Real challenges. By eliminating the need for extensive data collection, these datasets also foster fair comparisons and establish strong baselines for quadrotor research. VisFly supports common vision and motion sensors,  and is extensible to complex multi-camera systems.

One requirement in training flight policies is to query the nearby obstacles on the fly. 
Habitat-sim, specialized for ground robots, assumes that robots operate on a 2-dimensional surface. It precomputes the NavMesh, which projects spatial objects onto the floor and generates distance information to the closest obstacle. This approach is not suitable for aerial robots that move in 3D space. Therefore, we developed a spatial distance computation module via the CGAL\footnote{https://www.cgal.org/} package in C++, enabling it to detect the relative position of the nearest point on the obstacle without collision actually happening. 

VisFly supports parallelized agents as well as multi-agents within the same scene. These two modes share the same underlying framework. The key difference between them is that parallelized agents visually omit other drones and ignore collisions between them, while multi-agent environment includes interactions and collisions among all drones. 

\subsection{Modular design for Gym integration}
Gym \cite{brockman2016openai}  is an open-source toolkit providing a standard interface for reinforcement learning. It offers a collection of pre-built environments and a framework for custom environment development. Agents are able to interact with environments through \textbf{step()} function, obtaining rewards, observations, termination signals, and other information at each time step.

VisFly's ultimate mission is to facilitate the development of aerial robot learning for autonomous flight tasks. Therefore, we have taken the extensibility of the basic environment into account as much as possible when packaging the environment. Two abstract environment classes \textbf{droneGymEnvsBase}, \textbf{multiDroneGymEnvsBase} are provided. When users design custom tasks for a quadrotor, they need only define a class that inherits from the abstract gym environment class and overrides three functions: \textbf{get\_success()}, \textbf{get\_reward()}, and \textbf{get\_observation()} as shown in Lst. \ref{alg:inherit}. This approach creates a new gym environment tailored to the specific mission. We provide three examples in \refsec{Applications} for users to reference. This hierarchical architecture enables the training of models that can act as backbones for more complex tasks. For instance, a model trained for stable hovering can serve as the low-level policy or backbone for higher-level autonomous tasks such as flying in cluttered environments. 

% \begin{figure}[h]
% \centering
% \includegraphics[width=0.85\linewidth]{fig/inherit.png}
% \caption{High-level tasks definition by inheriting base class.}
% \label{fig:inherit}
% \end{figure}

\begin{lstlisting}[caption={Mocular class design for different tasks.    }, label={alg:inherit}]
class DroneGymEnvsBase:
    @abstract method
    def get_reward()
        return reward
    @abstract method
    def get_observation()
        return observation
    @abstract method
    def get_success()
        return is_success
    
class DroneNavigationEnvs(DroneGymEnvsBase)
    def get_reward()
        ... 
        return r_distance+r_speed-r_collision
    def get_observation()
        ... 
        return state+vision+target
    def get_sucess()
        ... 
        return status_reach_the_target
        
...
\end{lstlisting}

\subsection{Domain randomization for sensor data and flight states}
The gap between simulation and reality usually hinders the real-world deployment of models trained in simulation. In order to reduce the sim-to-real domain gap, one effective approach is to improve the realism of simulation, whose drawback is discussed in Introduction. The other approach is to enlarge the data distribution generated by simulators, strengthening the in-distribution generalization of models, which is called domain randomization\cite{tobin2017domain}. For training vision-based flight policies, it is important to add noises to both the motion and vision sensors as listed in \reftab{tab:noise}.

\begin{table}[ht]
\scriptsize
\centering
\caption{Noise model of sensors\label{tab:noise}}
\label{tab:your-table}
\begin{tabularx}{\linewidth}{c | c}
\toprule
\textbf{Sensor} & \textbf{Noise Model}  \\
\midrule
{IMU} & Normal  \\
{RGB} & Normal, Poisson, Salt\&Pepper, Speckle \\
{Depth} & Normal, Poisson, Salt\&Pepper, Speckle, RedWood\cite{choi2015robust}    \\
{Segmentation} &  Normal, Poisson, Salt\&Pepper, Speckle \\
\bottomrule
\end{tabularx}
\end{table}
In addition to adding noises to sensors,  VisFly allows for the randomization of initial conditions, including position, velocity, orientation, and angular velocity, which can follow either a normal or uniform distribution. To achieve better generalization of the policy, agents are reset with initial positions away from obstacles in cluttered environments. An important contribution of VisFly is its capability to run multiple scenes distributively and aggregate observations centrally. During training, scenes are treated as datasets, allowing them to be sampled and shuffled. To the best of our knowledge, VisFly is the first simulator that supports simultaneous execution of multiple scenes for enhanced scene randomization. 

\begin{figure}[h]
\centering
\includegraphics[width=\linewidth]{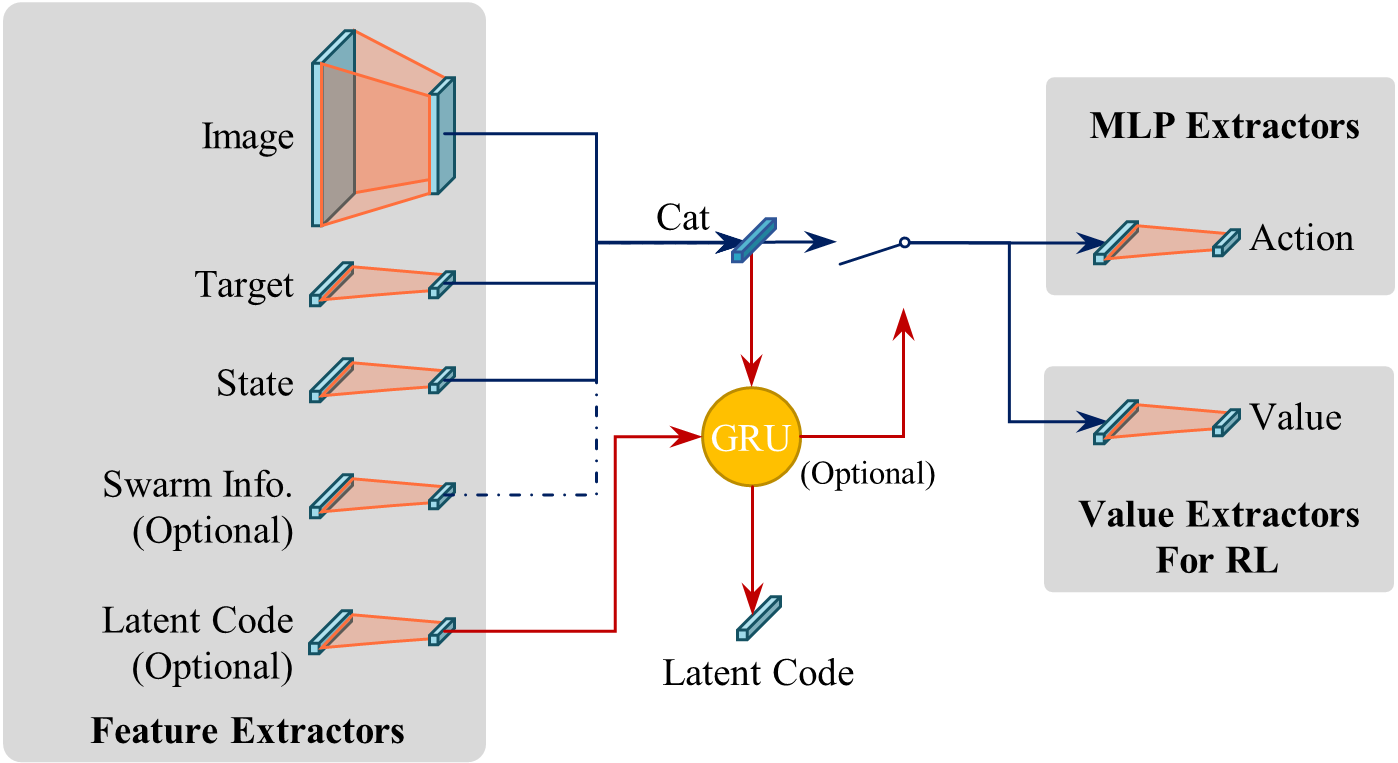}
\caption{Neural network architecture used across different tasks. The image feature extractor has been modified to be more customizable, including backbones such as ResNet\cite{he2016deep}, MobileNet\cite{sandler2018mobilenetv2}, EfficientNet\cite{tan2019efficientnet}. The policies for the three tasks follow this architecture but are slightly different. VisFly additionally incorporates a recurrent network interface \cite{chung2014empirical}. Detailed policy setting is introduced in VisFly homepage.
}
\label{fig:netArchitecture}
\end{figure}

\begin{figure*}[h!]
\centering
\includegraphics[width=\textwidth]{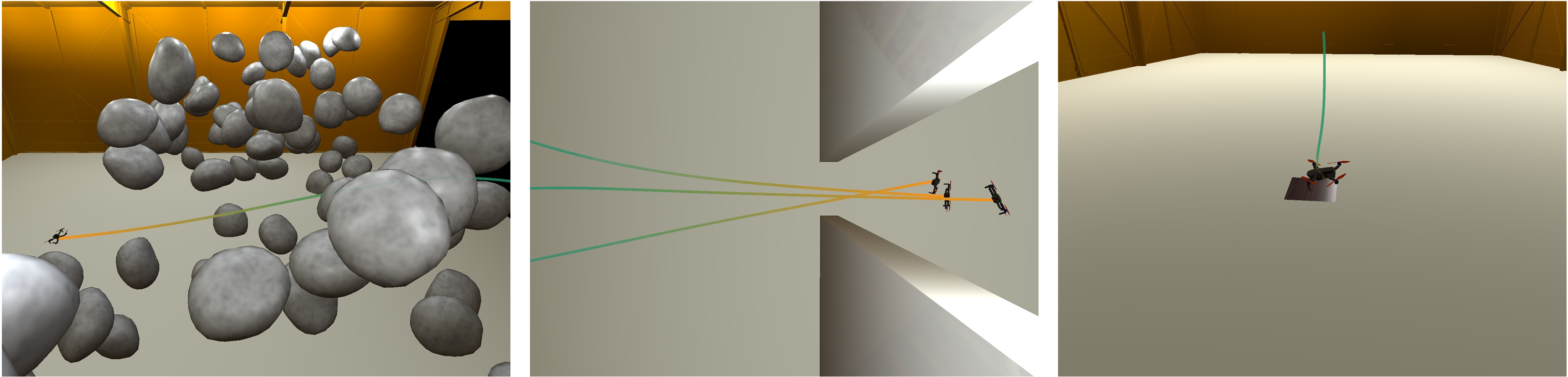}
\caption{Visualization of \textbf{Left}: navigation, \textbf{Middle}: cooperatively crossing, \textbf{Right}: landing tasks. }
% \caption{RGB, depth, and semantic segmentation captured in open-source and customized scenes. On the right is a swarm of quadrotors in a clear garage. testtestssdsdsdsdset}
\label{fig:example}
\end{figure*}

\section{Training examples}
\label{Applications}
To validate this simulator, we present three cases of training vision-based flight policies: (1) learning to land, (2) learning to navigate in a cluttered environment,  and (3) learning to cross the narrow gap cooperatively. 
% These representative tasks both need real-time visual information during flight. In traditional autonomous flight, the decision-making procedure has been divided into three components: perception, planning, and controlling. This modular architecture of perception, planning, and control in quadrotor autonomy enhances flexibility, maintainability, reliability, and efficiency by allowing independent development, testing, and optimization of each component. However, this approach introduces redundancy into the system, leading to potential decay or increased hardware requirements. To address these issues, the concept of learning from observations to actions, known as end2end learning, was proposed. end2end learning uses neural networks to replace multi-step procedures, significantly simplifying the entire system.

We use reinforcement learning to train end-to-end models for controlling the drones using vision inputs. To focus on training efficiency rather than the algorithms themselves, we employ PPO \cite{schulman2017proximal} from stable-baselines3\cite{raffin2021stable} (SB3) for its outstanding capability to handle high-dimensional observations. %, to train the models. 
The backbone architecture of the policy network follows the standard design in SB3, as shown in \reffig{fig:netArchitecture}, with modifications to enhance its adaptability across various tasks.  
% The image feature extractor is a conventional neural network that comprises three layers. Each layer includes a convolutional layer, an activation function, and a max pooling operation, followed by a linear layer with 128 units. The convolutional layers have kernel sizes of 5, 3, and 3, with channel counts of 6, 12, and 18. Each max pooling operation uses a kernel size of 2. The target extractor, state extractor, and swarm extractor are all multilayer perceptrons (MLPs) with two linear layers, having 128 and 64 units. The action net and value net are also MLPs, with two linear layers of 64 units. All activation functions are ReLU. This architecture has been parameterized, making it convenient to fine-tune.
The action type in all the examples is CTBR. The quadrotor takes its full state (position, orientation, linear velocity, angular velocity), visual image, and target as observations. Most of the hyper-parameters of PPO are set to default values. Detailed settings, the network architecture, the reward, done, and observation functions are provided in the source code. 
\begin{figure}[h]
    \centering
    \includegraphics[width=\linewidth]{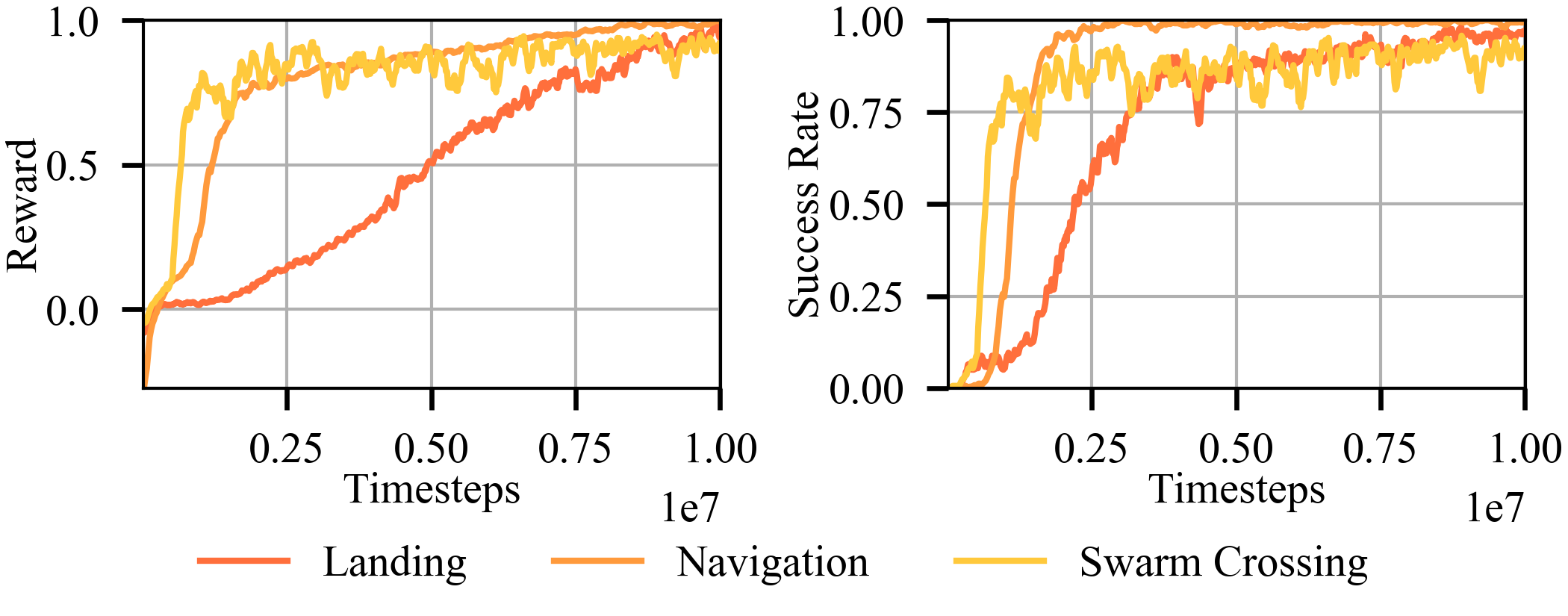}
    \caption{Reward and success rate curves during training of navigation, corporately crossing, landing task.}
    \label{fig:train_curve}
\end{figure}
\subsection{Learning to navigate in a cluttered environment}
\label{Navigation}
In this task, the quadrotor must reach the target position safely while avoiding obstacles during flight. It is one of the most fundamental tasks for quadrotors in real-world applications. We set a garage scene filled with random stones of the density of $0.15/m^3$ in the spatial space as shown in \reffig{fig:example}. This scenario is more challenging than dense scenes with trees or columns because it also requires the quadrotor to account for vertical motion. The quadrotor starts from an initial position, navigates through the stone-filled region, and eventually reaches the target position. It uses a $64\times64$ depth image as the visual observation and the final position as the target observation. Training takes 0.73 hours. To our knowledge, this is the first instance of end-to-end autonomous navigation in a cluttered environment without employing training techniques such as curriculum learning, privileged information, or customized primitives within the simulation.

\subsection{Learning to cross a narrow gap cooperatively}
We inserted two walls in the clear garage to create a narrow gap between them as \reffig{fig:example}. Three quadrotors start from one side of the walls, fly towards distributed targets on the other side, and pass through the narrow gap simultaneously. They must cooperate by considering each other's state to avoid obstacles and other agents safely. Each quadrotor uses a $64\times64$ depth image as its visual observation, the final position as its target observation, and the states of other agents as its group member observation. Unlike single-agent tasks, VisFly supports only one swarm per scene, which limits the maximum batch size due to memory constraints. This training takes 1.1 hours. The results show that by observing the states of other agents and the environment, the quadrotors can effectively fly across the gap and avoid collisions with each other.
\subsection{Learning to land}
In this task, the quadrotor starts in flight, gradually decelerates, and eventually lands safely on the landing area. It continuously monitors the limited landing area directly below, which is represented as a black rectangle positioned horizontally on the ground, as shown in \reffig{fig:example}. The quadrotor calculates the center of the $0.5\times0.5m^2$ black rectangle in its field of view via threshold segmentation, strives to keep it centered, and incorporates it into its observation. During the flight, the quadrotor captures only $64\times64$ RGB image from the camera rather than acquiring precise spatial information about the landing area. This training process takes 1.7 hours, whose extra time (compared to Navigation) accounts for computing the center of the landing area.
% \subsection{Recurrent Policy for Navigation \textbf{(DELETE OR RESERVE)}}
% Most of RL algorithms and instances are based on the assumption of Markov Decision Process (MDP). MDP assumpts that agents have access to entire observation, and current decision is irrelevant with the sequence of previous or further observation. Nevertheless, the real process can only obtain part of environments, also called as Partially Observable Markov Decision Process (POMDP), just like \refsec{Navigation} task, agents cannot see obstacles hidden behind other objects. For high-level tasks like searching, the explored area and previous trajectory should be take into consideration as well. Recurrent policy (RP) has made progress implemented in low-dimension problems \cite{kapturowski2018recurrent}. Therefore, we have reserved an interface for the transition to RP and conduct a preliminary study on the effectiveness in robotics like \refsec{Navigation} task. As \reffig{fig:recurrent}, the reward and success rate curves of RP increases faster a bit than the other one. This is probably that RP records the previous seen hidden obstacle from different perspectives. This phenomenon is going to be analyzed in the following research. 

% \begin{figure}[h]
% \centering
% \includegraphics[width=\linewidth]{fig/RP_train_res.png}
% \caption{Training results of navigation with recurrent policy and common policy. }
% \label{fig:recurrent}
% \end{figure}

\section{Conclusion And Discussion}
This paper presents VisFly, a quadrotor simulator designed for efficient training of vision-based flight policies. Leveraging differentiable physics and Habitat-Sim's rendering capabilities, VisFly supports single and multi-agent simulations across various controller types. Integrating with the gym environment provides a standardized interface for various reinforcement learning algorithms. Experimental results on landing, cluttered environment navigation, and cooperative gap crossing demonstrate the simulator's effectiveness. VisFly serves as a robust foundation for future research, offering baseline implementations and extensibility for advanced quadrotor tasks.

VisFly does not build in high-level controllers like Model Predictive Control (MPC) because of its requirements for substantial computational resources. Currently research \cite{song2023reaching} has demonstrated that these basic controllers are enough to train autonomous policies. In future work, we plan to integrate real-world dynamics into VisFly to enable Hardware-in-the-Loop (HITL) capabilities, and keep updating it.

%%%%%%%%%%%%%%%%%%%%%%%%%%%%%%%%%%%%%%%%%%%%%%%%%%%%%%%%%%%%%%%%%%%%%%%%%%%%%%%%



% \section*{ACKNOWLEDGMENT}

% The preferred spelling of the word ÒacknowledgmentÓ in America is without an ÒeÓ after the ÒgÓ. Avoid the stilted expression, ÒOne of us (R. B. G.) thanks . . .Ó  Instead, try ÒR. B. G. thanksÓ. Put sponsor acknowledgments in the unnumbered footnote on the first page.

%%%%%%%%%%%%%%%%%%%%%%%%%%%%%%%%%%%%%%%%%%%%%%%%%%%%%%%%%%%%%%%%%%%%%%%%%%%%%%%%

% References are important to the reader; therefore, each citation must be complete and correct. If at all possible, references should be commonly available publications.
\newpage
\bibliographystyle{IEEEtran}
\bibliography{ref.bib}

\end{document}